\documentclass[runningheads,a4paper]{llncs}

\usepackage{amssymb}
\usepackage{graphicx}
\usepackage{array}
\usepackage[utf8]{inputenc}
\usepackage{url}
\urldef{\mailsa}\path|{alfred.hofmann, ursula.barth, ingrid.haas, frank.holzwarth,|
\urldef{\mailsb}\path|anna.kramer, leonie.kunz, christine.reiss, nicole.sator,|
\urldef{\mailsc}\path|erika.siebert-cole, peter.strasser, lncs}@springer.com|
\newcommand{\keywords}[1]{\par\addvspace\baselineskip
\noindent\keywordname\enspace\ignorespaces#1}

\begin{document}

\mainmatter  

\title{Neural Machine Translation with\\Recurrent Highway Networks\\}

\titlerunning{Neural Machine Translation with Recurrent Highway Networks}

%
%
\author{Maulik Parmar\inst{1}%
\and\\
V.Susheela Devi\inst{2}}
%

\institute{Indian Institute of Science,\\
Bangalore-560012, Karnataka, India\\
}

%
%

\toctitle{Lecture Notes in Computer Science}
\tocauthor{Authors' Instructions}
\maketitle

\begin{abstract}
\emph{Recurrent Neural Networks have lately gained a lot of popularity in language modelling tasks, especially in neural machine translation(NMT). Very recent NMT models are based on Encoder-Decoder, where a deep LSTM based encoder is used to project the source sentence to a fixed dimensional vector and then another deep LSTM decodes the target sentence from the vector. However there has been  very little work on exploring architectures that have more than one layer in space(i.e. in each time step). This paper examines the effectiveness of the simple Recurrent Highway Networks(RHN) in NMT tasks. The model uses Recurrent Highway Neural Network in encoder and decoder, with attention .We also explore the reconstructor model to improve adequacy. We demonstrate the effectiveness of all three approaches on the IWSLT English-Vietnamese dataset. We see that RHN performs on par with LSTM based models and even better in some cases.We see that deep RHN models are easy to train compared to deep LSTM based models because of highway connections. The paper also investigates the effects of increasing recurrent depth in each time step.} 
\end{abstract}

\section{Introduction}
Neural Machine Translation(NMT) is a recent approach towards machine translation( \cite{6}, \cite{2},  \cite{1} ,\cite{3}).In contradiction with the conventional Statistical Machine Translation(SMT) system  ( \cite{7}) which consists of many small sub-components that are tuned separately, in neural machine translation the whole neural network is jointly trained to maximize the conditional probability of a correct translation given a source sentence, using the bilingual corpus.\\
The use of Neural Networks for Machine Translation leads to fluent translation.Popular neural machine translation models(\cite{2}, \cite{1}, \cite{3}) use stacked LSTMs, which read through the source sentence one word at a time till it reaches the end of sequence symbol$<eos>$ tag, then it starts to emit target words one by one till it generates the end of sequence $<eos>$ tag.\\
We move in the same direction, but instead of stacking layers of LSTM, we stack layers within each time step, as illustrated in Figure 1. Each time block is recurrent highway network following the work of \cite{4} . With this architecture we demonstrate their effect on machine translation when layers within a time step is increased one by one.Parallelly with the neural machine translation(NMT), the concept of "attention" has gained a lot of popularity in recent years( \cite{2}, \cite{3}). We follow the attention model by  \cite{3} as our base model. With IWSLT English-Vietnamese dataset we found that without attention increasing the recurrent depth does not affect the {BLEU}\cite{8} score much, whereas with attention it shows a significant increase in the BLEU score with increasing recurrent depth.

    \begin{figure*}
    \begin{center}
    \centerline{\includegraphics[width = \textwidth]{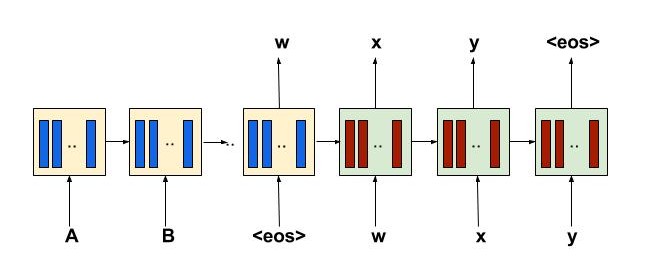}}
    \caption{Neural machine translation with recurrent highway architecture for translating a source sequence A B into  a  target  sequence X Y.(Proposed Model without attention)}
    \label{icml-historical}
    \end{center}
    \vskip -0.3in
    \end{figure*}

The problem with these vanilla attention models is that translation generated by these models often lack adequacy.There is a problem of over-translation and under-translation i.e some words are translated more than once ,while some words are not translated at all.To address this problem we use reconstructor model(\cite{9}) which we will discuss in Section 4.

\section{Related Work}    
 Even prior to the recent end-to-end NMT models neural networks were used for SMT based translation with some success.The concept of end-to-end learning for machine translation has been attempted in the past with limited success. From then its a long journey to the end-to-end models(\cite{1,2,3,6}) which have outperformed the previous SMT baselines(\cite{7}).\\
    Our work being the first application of RHN in neural machine translation task, is very close to the work  \cite{1},\cite{3} and  \cite{9}. At the same time we take the motivation to apply the RHN in this area from \cite{5}; who has shown the success of recurrent highway network(RHN) in automatic speech recognition.

\section{Recurrent Highway Network}
    Recurrent state transition in RNN is described by  $y^{[t]} = f(Wx^{[t]} + Ry^{[t-1]} + b)$. Similarly, a Recurrent Highway Network(RHN) layer has multiple highway layers in the recurrent state transition.RHN were first introduced in \cite{4}.\\
    \\
    Let, $W_{H,T,C}\in R^{n\times m}$ and $R_{H_l,T_l,C_l}\in R^{n\times n}$ represent  the weights matrices of the H nonlinear transform and the T and C gates at layer $l={1,...,L}$. The biases are denoted by $b_{H_l,T_l,C_l}\in R^{n}$ and let $s_l$ denote a intermediate output at layer $l$ with ${s{_{^{[t_0]}}}} = {y^{[t-1]}}$. Then an RHN layer with a recurrence depth of L is described by:
    \begin{equation}
    {s_l}^{[t]} = {h_l}^{[t]}*{s_l}^{[t]} + {s_{l-1}}^{[t]}*{c_l}^{[t]}
    \end{equation}
    where
    \begin{eqnarray}
    {h_l}^{[t]}& = tanh(Wx^{[t]}I_{[l=1]} + R_{H_l}*{s_l}^{[t]} + b_{H_l})\\
    {\textbf{t}_l}^{[t]}& = \sigma(Wx^{[t]}I_{[l=1]} + R_{T_l}*{s_l}^{[t]} + b_{T_l})\\
    {c_l}^{[t]}& = tanh(Wx^{[t]}I_{[l=1]} + R_{C_l}*{s_l}^{[t]} + b_{C_l})
    \end{eqnarray}
    
    and $I_{[l=1]}$ is the indicator function.\\

    It is to be observed that a RHN layer with $L = 1$ is essentially a basic variant of an LSTM layer. The computation graph for a RHN block is illustrated in Figure 2. 
    \begin{figure}[h]
    \begin{center}
    \centerline{\includegraphics[width= \columnwidth]{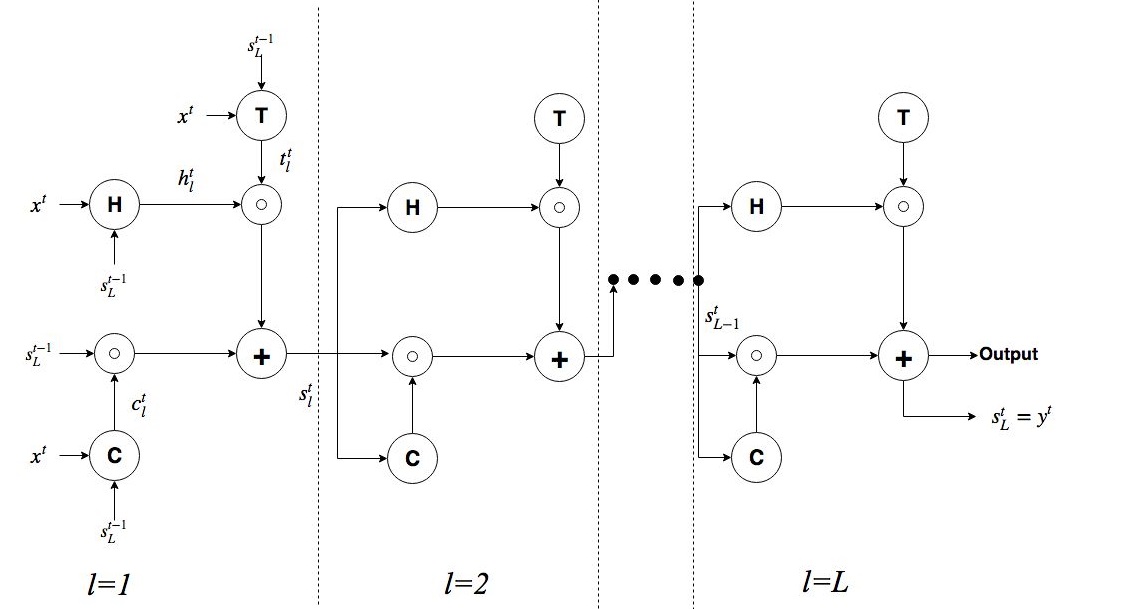}}
    \caption{Schematic showing computation within an RHN layer inside the recurrent loop. Vertical dashed lines delimit
stacked Highway layers. Horizontal dashed lines imply the extension of the recurrence depth by stacking further layers.
$H,T,C$ are the transformations described in equations 2, 3 and 4, respectively. (\cite{4})}
    \label{icml-historical}
    \end{center}
    \vskip -0.3in
    \end{figure}
    
\section{Model}
Many sequential processing tasks require complex
nonlinear transition functions from one step
to the next. However, recurrent neural networks
with "deep" transition functions remain difficult
to train, even when using Long Short-Term Memory
(LSTM) networks as shown by (\cite{4},\cite{11}).We use existing neural machine translation models with recurrent highway networks(which extend the LSTM architecture to allow
step-to-step transition depths larger than one) as their recurrent block .
\subsection{Model 1}
    We propose a neural machine translation model using the recurrent highway network architecture given by \cite{4}, and we use encoder-decoder system on this architecture with attention as given by \cite{3} as our two base models.We use RHN instead of LSTM to model conditional probability of target sentence given source.\\
    
\subsection{Model 2}
The main problem with previous models and attention mechanisms was that sometimes translation produced by them lacked adequacy.There are two reasons for this:
\begin{itemize}
\item {\textbf{Over-Translation-}}
     Some words are translated again and again.This is because there is no mechanism in the previous architecture to see that a word which is translated is not heavily attended.
\end{itemize}
\begin{itemize}
\item {\textbf{Under-Translation-}}
     Some words are not translated at all.This is because there is no mechanism to make sure that all words are attended or translated.
\end{itemize}
\begin{figure*}[h]
    \begin{center}
    \centerline{\includegraphics[width =\columnwidth]{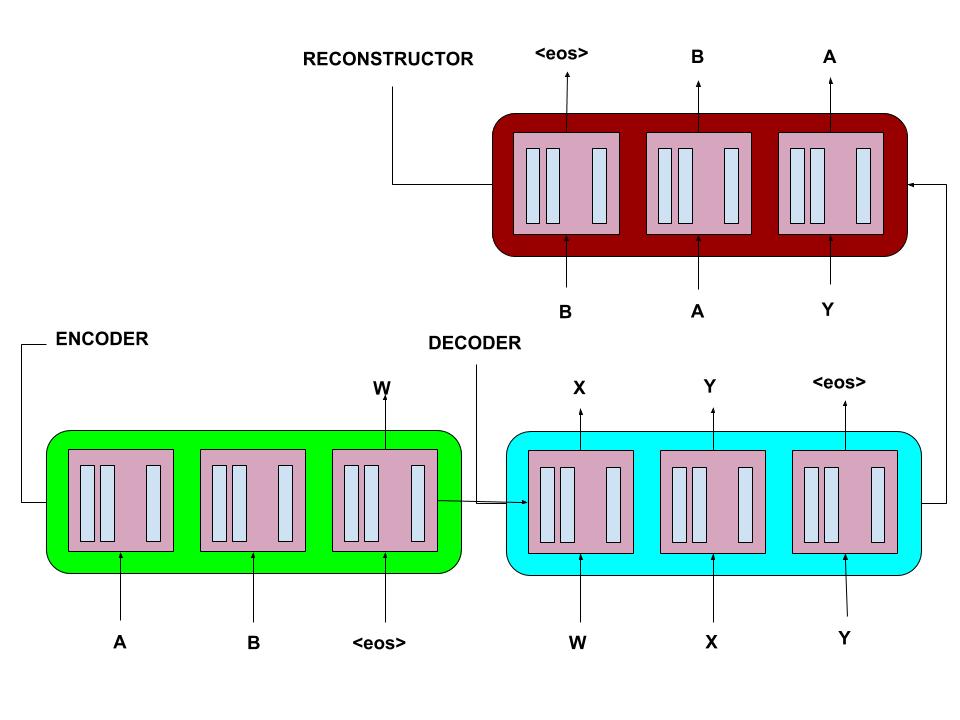}}
    \caption{Example of an reconstructor-based NMT system as described in ( \cite{9}).Green blocks are part of reconstructor. For clarity, the embedding and projection layers are not shown in Figure.}
    \label{icml-historical}
    \end{center}
    \vskip -0.3in
    \end{figure*}

To tackle this problem we use model 1 with a reconstructor block as in \cite{9} .The encoder-decoder framework is used to translate source sentence into target sentence.The decoder-reconstructor framework is used to convert target sentence into source sentence again.There is maximum information flow from encoder to decoder so that the reconstructor is able to reproduce source sentence from target sentence.By using reconstructor,model 2 tries to make sure that almost all source words are translated into target sentence so that reconstructor can reconstruct source words later.Thus ,translation produced is fluent and adequate.We use RHN instead of LSTM to model conditional probability of target sentence given source and to model conditional probability of source sentence given target hidden states (in reconstructor block).    
The thing to be noted is that we use the reconstructor block only during training phase.During inference,we use the regular encoder-decoder framework without reconstructor block.Unlike \cite{9},we only use reconstructor in training phase and not during inference.The loss function used for training comprises of two terms,decoder loss and reconstructor loss.
We train the neural network by minimizing two loss terms.The first term decoder loss $L_d$,is the log probability of a correct translation T given the source sentence S.
So the training objective is
\begin{equation}
        L_d =\frac{-1}{\left \| S \right \|}\sum_{(T,S)\in S}\log p(T/S)
    \end{equation} 
The second term reconstructor loss $L_r$,is the log probability of a correct translation S given target hidden states H.    
\begin{equation}
        L_r =\frac{-1}{\left \| S \right \|}\sum_{(S,T)\in S}\log p(S/H)
    \end{equation}     

where S is the training set.
The final loss function is given by:
\begin{equation}
    L=L_d+\beta *L_r
\end{equation}

where $\beta$ is a hyperparameter which reflects the importance of reconstruction compared to translation task and should be tuned based on the dataset used.  
\section{Experiments}
    
    We applied our method on the IWSLT English-Vietnamese dataset in both directions. Reference sentences are used to compute BLEU score of the translated sentences. We changed the number of hidden units per cell and recurrence depth(no. of layer in each time step) while keeping all other parameters constant. We report the performance of these translation models in Table 1 , and visualize the resulting translations.  We report the performance of these translation models  using LSTM in Table 2.
    
        \subsection{Dataset details}
        IWSLT English-Vietnamese data set is used for our experiment. The dataset has $133k$ sentence pairs of training sentences. We used 'tst2012'(en|vi) and 'tst2013'(en|vi) as our development and test dataset respectively. The dataset is available at Stanford NLP web page.Out of vocab words produced during translation, are replaced by an '$<unk>$' token.
        \subsection{Evaluation Metrics}
        \begin{itemize}
        \item Perplexity :In natural language processing, perplexity is a way of evaluating language models. A language model is a probability distribution over entire sentences or texts.
        \end{itemize}
        \begin{equation}
         Perplexity=2^{-\frac{1}{N}\sum_{n=1}^N\sum_{t=1}^Tln p_{target} }
        \end{equation}
        \begin{itemize}
        \item BLEU Score:{BLEU}\cite{8}, or the Bilingual Evaluation Understudy, is a score for comparing a candidate translation of text to one or more reference translations.
        \begin{equation}
         BLEU = min(1,\frac{output\_length}{reference\_length})\sqrt[4]{(\prod_{i=1}^4 precision_i)}
        \end{equation}
        \end{itemize}  
        
        \subsection{Training details}
        
        \begin{itemize}
            \item We have used SGD with a learning rate of $0.1$ .
            \item We have used a dropout of $20\%$.
            \item A batch size of 32 is used.
            \item Roughly all the experiments ran for 30 epochs.
        \end{itemize}

    \begin{table}
    \centering
    \label{my-label}
     \caption{Test perplexity and BLEU score obtained on tst2013 test set with RHN as recurrent network.Source -English, Target-Vietnamese. }
   
    \begin{tabular}{|m{4cm}|c|c|c|c|c|}
    \hline
    \multicolumn{1}{|l|}{Model} & \multicolumn{1}{|l|}{depth} & \multicolumn{1}{|l|}{Layers} & \multicolumn{1}{|l|}{hidden units} & \multicolumn{1}{l|}{Test perplexity} & \multicolumn{1}{l|}{Test BLEU} \\ \hline
    Model1 and greedy search  & 3 & 1  & 128 & 15.82 & 21.8 \\ \hline
    Model1 and greedy search  & 3 & 1 & 256 & 12.56 & 23.1 \\ \hline
    Model2 with $\beta=1$ and greedy search  & 2 & 2 & 128 & 14.24 & 22.0 \\ \hline
    Model2 with $\beta=1$ and greedy search  & 2 & 2 &  256 & 14.98 & 22.6 \\ \hline
    Model2 with $\beta=1$ and greedy search & 2 & 2 &  512 & 14.97 & 22.9 \\ \hline
    Model2 with $\beta=0.5$ and greedy search  & 2 & 2 &  128 & 14.38 & 22.8 \\ \hline
    Model2 with $\beta=0.5$ and greedy search  & 2 & 2 &  256 & 12.46 & 23.4 \\ \hline
    Model2 with $\beta=0.5$ and greedy search  & 2 & 2 &  512 & 15.46 & 22.5 \\ \hline
    Model2 with $\beta=0.1$ and greedy search & 2 & 2 &  256 & 12.41 & 24.0 \\ \hline
    Model2 with $\beta=0.1$ and beamsearch (bw=10) & 2 & 2 &  256 & 11.98 & 24.9 \\ \hline
    Model2 $\beta=0.1$ and greedy search & 7 & 1 & 256 & 12.29 & 23.9 \\ \hline
    Luong attention model with LSTM as in paper ( \cite{10})& - & - & - & - & 23.3 \\ \hline

    \end{tabular}
    \end{table}
    \begin{table}
    \centering
    \label{my-label}
     \caption{Test perplexity and BLEU score obtained on tst2013 test set with LSTM as recurrent network,$\beta=0.1$ and 256 hidden units.Source -English, Target-Vietnamese. }
    \begin{tabular}{|m{8cm}|c|c|}
    \hline
    \multicolumn{1}{|l|}{Model} & \multicolumn{1}{l|}{Test perplexity} & \multicolumn{1}{l|}{Test BLEU} \\ \hline
    Reconstructor model,  with greedy search(\cite{9})  & 13.31 & 22.3 \\ \hline
    Reconstructor model,with beam search(bw=10)(\cite{9})  & 11.45 & 23.4 \\ \hline
    
    Attention model,layers=1 with greedy search & 13.52 & 22.1 \\ \hline
    Attention model,layers=2 with greedy search & 12.07 & 23.1 \\ \hline
    
    \end{tabular}
    \end{table}

    \subsection{Sample Translations}
\begin{table*}
\centering
\label{my-label}
\caption{Translation of randomly picked sentences with Model 1 and Model 2.}
\begin{tabular}{|c|l|}
\hline
Ref & \begin{tabular}[c]{@{}l@{}}I even went through an identity crisis .\end{tabular} \\ \cline{2-2} 
NMT1 & \begin{tabular}[c]{@{}l@{}}I \&apos;ve experienced a crisis crisis of my origins .\end{tabular} \\ \cline{2-2} 
NMT2 & \begin{tabular}[c]{@{}l@{}}I experienced the crisis of my origins .
\end{tabular} \\ \hline
& \\ \hline
Ref & \begin{tabular}[c]{@{}l@{}}Where am I from ? Who am I ?\end{tabular} \\ \cline{2-2} 
NMT1 & \begin{tabular}[c]{@{}l@{}}Where do I come from ? And I m one person ?\end{tabular} \\ \cline{2-2} 
NMT2 & \begin{tabular}[c]{@{}l@{}}Where am I coming from ? Who am I ?\end{tabular} \\ \hline
& \\ \hline
\multicolumn{1}{|l|}{Ref} & Ông là ông ca tôi . \\ \cline{2-2}
\multicolumn{1}{|l|}{NMT1} & Ông là  tôi . \\ \cline{2-2} 
\multicolumn{1}{|l|}{NMT2} & Ông là ông tôi .\\ \hline
\end{tabular}

\end{table*}

\begin{table*}
\centering
\label{my-label}
\caption{Translation of randomly picked sentences with 128 hidden units and recurrence depth 3 and 4 without reconstructor.}
\begin{tabular}{|c|l|}
\hline
Ref & \begin{tabular}[c]{@{}l@{}}When I was little , I thought my country was the best on the planet ,\\ and I grew up singing a song called "Nothing To Envy".\end{tabular} \\ \cline{2-2} 
Depth-3 & \begin{tabular}[c]{@{}l@{}}When I was a little bit , I thought , "Can see this country is the best country\\ in the world and  I often sing " what I can't envy".\end{tabular} \\ \cline{2-2} 
Depth-4 & \begin{tabular}[c]{@{}l@{}}When I was a little boy , I think $<unk>$ is the best country\\ in the world and I often  sing "We doesn't have to envy".\end{tabular} \\ \hline
& \\ \hline
\multicolumn{1}{|l|}{Ref} & Since my family couldn't understand Chinese , I thought my family was going to be arrested . \\ \cline{2-2} 
\multicolumn{1}{|l|}{Depth-3} & Because my family didn't understand Mandarin , so I thought they would be arrested . \\ \cline{2-2} 
\multicolumn{1}{|l|}{Depth-4} & Because my family didn't know Chinese ,so I thought they would be arrested . \\ \hline
\end{tabular}
\end{table*}

    A few example translations produced by our model1 and model2  along with the reference sentences are shown in  Table 3.From this  randomly  picked  sample  it  is  visible  that, NMT2 avoids problems of over translation as seen in NMT1.Now we check the translation of the randomly picked sentences for recurrence depth 3 and 4 in Table 4. From this randomly picked sample it is visible that, increasing the recurrence depth helps to capture the semantics of the language better.                    
     
    \subsection{Model analysis}
    To check how good our model is learning with the change of layers i.e. recurrence depth, we plot the training perplexity of the model 1 with different number of recurrence depth.\\
    \\\\
    It can be seen that, there is a visible difference in the curve when we change depth from 2 to 3, and a bit less difference when we change from 3 to 4. Perplexity decreases with the increment of recurrence depth(Figure 4).As we can see in Table 5,RHNN models have less number of parameters as compared to LSTM  and are able to perform as good as LSTM models or in some cases even better.

    \begin{figure}[h]
    \begin{center}
    \centerline{\includegraphics[width= \columnwidth]{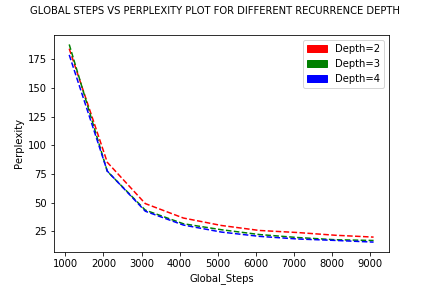}}
    \caption{Perplexity Vs. number of tensorflow steps.}
    \label{icml-historical}
    \end{center}
    \vskip -0.3in
    \end{figure}
    
    \begin{table}
    \centering
    \label{my-label}
    \caption{Number of parameters in reconstructor model.  }
    \begin{tabular}{|c|c|c|}
    \hline
    \multicolumn{1}{|l|}{Recurrent Unit} & \multicolumn{1}{l|}{No of hidden units} & \multicolumn{1}{l|}{Parameters(in millions)} \\ \hline
    LSTM with layers=2 & 128 & 1.2 \\ \hline
    RHNN with layers=1,depth=2& 128 & 0.6 \\ \hline
    LSTM with layers=2& 256 & 4.4 \\ \hline
    RHNN with layers=1,depth=2 & 256 & 2.1 \\ \hline
    RHNN with layers=2,depth=2 & 256 & 4.1 \\ \hline
    RHNN with layers=1,depth=7 & 256 & 2.1 \\ \hline
    
    \end{tabular}
    \end{table}

    \section{Conclusion}
    We use Recurrent Highway Networks, a powerful new model designed to take advantage of increased depth in the recurrent transition while retaining the ease of training of LSTMs.Our model works reasonably well with the IWSLT English-Vietnamese dataset in terms of BLEU score.The model shows significant amount of increase in score when we increase the recurrence depth  one layer at a time. The increasing trend is largely visible when we change from 2 to 3 layers, a bit less from 4 to 5 layers. It is almost negligible when we change the recurrence depth from 5 to 6. Hence, we claim in machine translation task, we get the advantage of increasing the recurrence depth , which is similar to the results as mentioned in {\cite{1}}; regarding the stack depth of LSTM's.
	Also to further produce adequate translation ,we use reconstructor model which also results in significant improvement.By using reconstructor block along with attention,we get improvement of almost 1 BLEU score compared to model without reconstruction.Also, we see that with beam search we were further able to improve translation and got improvement of around \textbf{1.5} BLEU score compared to benchmark result provided by\cite{10}.
	 \cite{4} showed that deep LSTM are difficult to train compared to deep RHN  .We can also see from Table 5 that the results are similar to work of \cite{4} which shows that in language modelling task also with almost half the parameters RHNN are able to outperform LSTM based language model.
	The same is true with neural machine translation task.


\end{document}